\documentclass[conference]{IEEEtran}
\IEEEoverridecommandlockouts
\usepackage{cite}
\usepackage{amsmath,amssymb,amsfonts}
\usepackage{algorithm,algorithmic}
\usepackage{graphicx}
\usepackage{graphics} 
\usepackage{threeparttable}
\usepackage{textcomp}
\usepackage{xcolor}
\usepackage{multirow}
\usepackage{mathptmx} 
\usepackage{times} 
\usepackage{scalerel}
\usepackage{epsfig} 
\usepackage{cancel}
\usepackage{pifont}
\usepackage{tabularx,booktabs}
\usepackage{hyperref}
\usepackage{enumitem}
\usepackage{booktabs}
\usepackage{color}
\usepackage{flushend}

\def\BibTeX{{\rm B\kern-.05em{\sc i\kern-.025em b}\kern-.08em
    T\kern-.1667em\lower.7ex\hbox{E}\kern-.125emX}}

\title{\LARGE \bf
Model-Free and Learning-Free Proprioceptive Humanoid Movement Control

\thanks{$^{\dagger}$ Department of Mechanical Engineering, National University of Singapore, Singapore {\tt\small borenjiang@u.nus.edu},\quad  {\tt\small mpegre@nus.edu.sg}}
\thanks{$^{*}$ Department of Mechanical Engineering, Johns Hopkins University, Baltimore, MD, USA {\tt\small hyf90916@gmail.com}
}

\author{Boren Jiang$^{\dagger}$, Ximeng Tao$^{\dagger}$, Yuanfeng Han$^{*}$, Wanze Li$^{\dagger}$, Gregory S. Chirikjian$^{\dagger}$}
}
\begin{document}

\maketitle
\thispagestyle{empty}
\pagestyle{empty}
\begin{abstract}
This paper presents a novel model-free method for humanoid-robot quasi-static movement control. Traditional model-based methods often require precise robot model parameters. Additionally, existing learning-based frameworks often train the policy in simulation environments, thereby indirectly relying on a model. In contrast, we propose a proprioceptive framework based only on sensory outputs. It does not require prior knowledge of a robot's kinematic model or inertial parameters. Our method consists of three steps: 1. Planning different pairs of center of pressure (CoP) and foot position objectives within a single cycle. 2. Searching around the current configuration by slightly moving the robot's leg joints back and forth while recording the sensor measurements of its CoP and foot positions. 3. Updating the robot motion with an optimization algorithm until all objectives are achieved. We demonstrate our approach on a NAO humanoid robot platform. Experiment results show that it can successfully generate stable robot motions.
\end{abstract}

 \section{Introduction}
Many works have addressed the topic of humanoid robot movement control over the last few decades. Great challenges in the planning and control of biped robots include high degrees of freedom (DoFs), nonlinear dynamics, and balancing with one foot on the ground. In the existing literature, model-based walking methods use either a simplified robot model \cite{kajita2014introduction} or a whole-body dynamic modeling approach \cite{hirai1998development,huang1999high,yamaguchi1999development}. Both rely on the accuracy of the robot's inertial parameters, such as mass, center of mass (CoM), and link lengths. Complex control methods are also needed to compensate for modeling errors. The remaining literature on walking focuses primarily on reinforcement learning (RL). However, these RL methods often require a robot's kinematic model and inertial parameters to build simulation environments, or a pre-planned walking trajectory is used as a reference signal \cite{meyer2014machine, wen2015q, garcia2020teaching}. And certain robustness adjustments, such as domain randomization, are also needed for sim-to-real transfer \cite{li2021reinforcement}.

A widely used model-based method is proposed by Shuuji et al. \cite{kajita20013d,kajita2003biped}. They facilitate the sophisticated nonlinear biped dynamics system into a simple 3D Linear Inverted Pendulum Mode (LIPM). A preview controller is used to track zero-moment point (ZMP) trajectories. Because the walking pattern is produced from a specified ZMP trajectory, this method is also called ZMP-based walking pattern generation \cite{vukobratovic1972stability}. Different from using the CoM of the whole robot, the authors in \cite{ficht2023direct} utilize a five-mass distribution model to plan whole-body motion. The momentum vector has a linear connection to the robot joint speed vector. Thus, other research uses a reference linear and angular momentum to generate whole body motion, called Resolved Momentum Control \cite{kajita2003resolved, gong2021one}. Angular momentum is also used in Capture Point (CP) to prevent a legged robot from falling by taking N steps before stopping \cite{pratt2006capture,koolen2012capturability,pratt2012capturability}. The CP is calculated by extending the LIPM with a flywheel body. George et al. \cite{mesesan2019dynamic} combine CP and a passivity-based controller for a humanoid robot to walk over compliant terrain. Hybrid Zero Dynamics (HZD)\cite{westervelt2003hybrid,hereid2019rapid,sreenath2011compliant} is a full-order model-based method. It depends on linear input and output, which can obtain more natural robot walking dynamics. However, these methods rely on the high accuracy of dynamic modeling, which may not be precise, and hence need external controllers. They also require high-performance CPUs and are computationally time-consuming. Additionally, these algorithms have high hardware requirements for robots. Thus, they may not be appropriate for small-sized and low-cost humanoid robots.
\begin{figure*}[t!]
\centering
\includegraphics[width=0.9\textwidth]{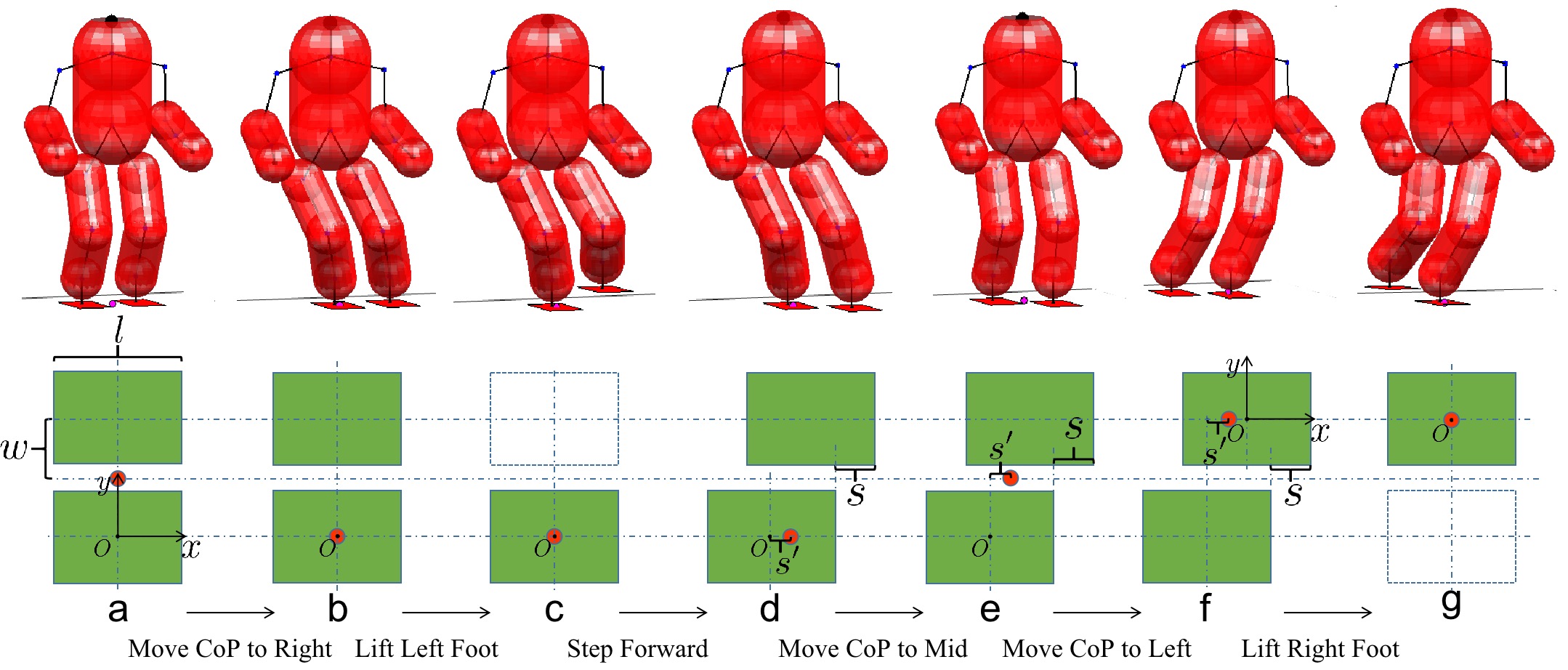}
\caption{Planning CoP transition and foot placement for humanoid robot walking pattern. The red point is the CoP of the robot. The floating base has its origin $O$ on the center of the foot, which is in the same side as the CoP. For example, in transition a $\rightarrow$ e (including bcd), the CoP is moving on the right side of the robot, so the origin is on the right side. The $z$-axis points up, and the $x$-axis points to the front of the robot.}
\label{gait}
\end{figure*}

RL-based methods also show great potential in robot locomotion without detailed model-based planning and control in real-time. Instead, they rely heavily on offline training. However, directly training the RL policy on a real-world robot can be risky and may cause damage. Therefore, researchers typically train in simulation environments. To bridge the gap between simulation and the real world, complex robustness regulations are required. In \cite{peng2017deeploco}, the authors demonstrate that a 3D biped can learn locomotion skills in simulation with limited prior model knowledge using a two-level hierarchical deep RL structure. Both levels use a uniform style of the actor-critic algorithm. Zhaoming et al. \cite{xie2018feedback} use a multi-layer neural network with policy-gradient to learn a feedback controller for robot walking in simulation. However, the method requires a pre-defined joint angle trajectory as a reference signal and acquires the robot's full state space information. But in reality, the states must be measured with noisy sensors. Combining with HZD, an RL framework is built to achieve variable speed controller learning for 3D robot walking in \cite{castillo2020hybrid}. Nevertheless, the method is only validated in a customized simulation environment and not on a real robot. Other works aim to train stable policies for robots walking in a complex environment. In \cite{krishna2021learning}, a gradient-free RL algorithm is used to achieve robust biped walking on varying sloped terrain in simulation. Diego et al. \cite{ferigo2021emergence} train a push-recovery whole-body motion planning policy in simulation with an RL framework. Although the works cited above obtain stable gait policies in simulation, the physics engines in the simulation environments still require robot dynamics models. That is, the above approaches cannot be generalized to robots without kinematic models or inertial parameters. 

In this paper, a model-free framework for humanoid robot movement control is introduced. As it is an initial effort in this area, only quasi-static situation is addressed here. This proprioceptive framework depends only on sensor data, which does not need any knowledge of robot parameters, such as inertial parameters and kinematic models. Thus, the difficulties induced by model uncertainty can be resolved. The approach can be directly implemented on real robots without pre-training in simulation or reference trajectories. Therefore, it has the potential to be extended to explore unknown environments. Compared to other frameworks, our method is much simpler, requiring only low computing power, data storage capacity, and graphics card performance.  Furthermore, the loss function can converge in only dozens of updates, indicating that a stable motion trajectory can be obtained in a short amount of time, without prior knowledge of a robot.



\section{Methodology}\label{Methodology}
In order to evaluate the effectiveness of our method, we conduct two experiments, one focus on walking and another on self-calibration of shoe sensors. In this section, we provide a detailed explanation of the walking experiment as an example to illustrate our approach.

To generate the motion trajectory for one walking cycle, several pairs of the CoP and foot positions are pre-designed. These pairs are treated as objectives for the robot to reach during its walking \ref{CoP Design}. We then define a cost function based on the Cartesian distances from the measured CoP and foot positions to their corresponding objectives \ref{Cost Function Design}. Then, the robot slightly moves each of its leg joints forward and backward, while the robot's CoP and foot positions are collected by sensors at each time \ref{Experiment Setup}. After this, the local derivatives can be approximated using the finite difference of sensor outputs without an explicit function. Finally, an optimization algorithm is used to update the robot's motion till all the objectives are achieved. \ref{optimization}

\subsection{CoP and Foot Objectives Design}\label{CoP Design}
This part introduces details of our objective CoP and foot position design. No prior knowledge of a robot's kinematic model or inertial parameters is used in this procedure.

The ZMP reference trajectory is designed by the humanoid robot's footsteps to stay in balance. The robot's slow movement can be approximated to quasi-static \cite{han2022watch}, in which case the ZMP and COP are equivalent. A simplified CoP path is used as shown in Fig.\ref{gait}. To begin, the robot is given the target of moving its CoP horizontally along the $y$-axis to the middle of the right foot while keeping both feet at the origin (Fig. \ref{gait}a $\rightarrow$ b). When the CoP is in the center of the right foot, the robot can lift its left foot vertically without falling, as Fig. \ref{gait}b $\rightarrow$ c. After the left foot is lifted to a certain height, the robot can step forward for a certain distance $s$ as Fig. \ref{gait}c $\rightarrow$ d. In the meantime, the CoP of the robot should also move forward for half of $s$, which makes sense that the CoP should stay close to the center of the support polygon \cite{erbatur2008natural}. Then, the robot moves its CoP horizontally along the $y$-axis back to the center as Fig. \ref{gait}d $ \rightarrow $ e. The solved joint trajectories are stored and the motion of the other leg can just mirror the stored configuration trajectories due to the symmetry of biped robots. Therefore, setup methods in transition Fig. \ref{gait}e $\rightarrow$ f $\rightarrow$ g are the same as above with the inverse of left and right. 

\subsection{Cost Function Design}\label{Cost Function Design}
The general formula of the cost function for each posture transition is
\begin{equation}
    f = \frac{1}{2}(||{\bf C}-{\bf C}_d||^2_{{\bf w_{c}}}+\sum ||{\bf P}[j]-{\bf P}_{d}[j] ||^2_{{\bf w_{p}}})
    \label{cost}
\end{equation}
where ${\bf C}=[x,y]^T$ is the measured CoP of the robot, ${\bf C}_d$ is the desired CoP, $j=l,r$ means left and right foot, ${\bf P} = [x,y,z,\alpha,\beta,\gamma]^T$ is the current position and Euler angle of the robot's foot, and ${\bf P}_d$ is the desired foot posture, $w_{c}$ and $w_{f}$ are the weights for the cost terms. Obviously, $f$ is a function of the robot joint angles, $\bf q$. As a result, one can find the derivatives of $f(\bf q)$ and minimize the cost. But in the present method, knowledge of $f({\bf q})$ as an analytical function is not required. The details will be shown in \ref{Map}. According to Fig. \ref{gait}, the desired terms in the cost for each walking pattern transition are in Table~\ref{t}.  Because the desired rotation of the robot's feet is always zero, only 3-D positions of feet are listed. 
\begin{table}[H]
\caption{Desired cost design} 
\centering 
\begin{threeparttable}
\begin{tabular}{|c|c|ccc|}
\hline
\multirow{2}{*}{Index} & \multirow{2}{*}{Gait Transition} & \multicolumn{3}{c|}{Desired Terms}                                                                       \\ \cline{3-5} 
                       &                                  & \multicolumn{1}{c|}{${{\bf C}_d}^T$}           & \multicolumn{1}{c|}{${{\bf P}_{d}[l]}^T$}                            & ${{\bf P}_{d}[r]}^T$              \\ \hline
1 & {\bf a} $\rightarrow$ {\bf b} & \multicolumn{1}{c|}{\multirow{2}{*}{$[0\;0]$}} & \multicolumn{1}{c|}{${[}0\;W\;0{]}$} & \multirow{4}{*}{${[}0\;0\;0{]}$} \\ \cline{1-2} \cline{4-4}
2                      & {\bf b} $\rightarrow$ {\bf c}                           & \multicolumn{1}{c|}{}             & \multicolumn{1}{c|}{${[}0\;W\;h{]}$}                  &                 \\ \cline{1-4}
3                      & {\bf c} $\rightarrow$ {\bf d}                           & \multicolumn{1}{c|}{$[s'\;0{]}$}  & \multicolumn{1}{c|}{\multirow{2}{*}{${[}s\;W\;0{]}$}} &                 \\ \cline{1-3}
4                      & {\bf d} $\rightarrow$ {\bf e}                           & \multicolumn{1}{c|}{$[s'\;w]$}  & \multicolumn{1}{c|}{}                              &                 \\ \hline
5                      & {\bf e} $\rightarrow$ {\bf f}                           & \multicolumn{1}{c|}{$-[s'\;0{]}$} & \multicolumn{1}{c|}{${[}0\ 0\ 0{]}$}                   & $-[s\; W\;0{]}$ \\ \hline
\end{tabular}
\end{threeparttable}
\label{t}
 \begin{tablenotes}
\footnotesize
\item where $s'=s/2$, $W=2w$
\end{tablenotes}
\end{table}
\subsection{CoP and Foot Positions Measurement}\label{Experiment Setup}
\begin{figure}[t!]
\centering
\includegraphics[width=0.95\linewidth]{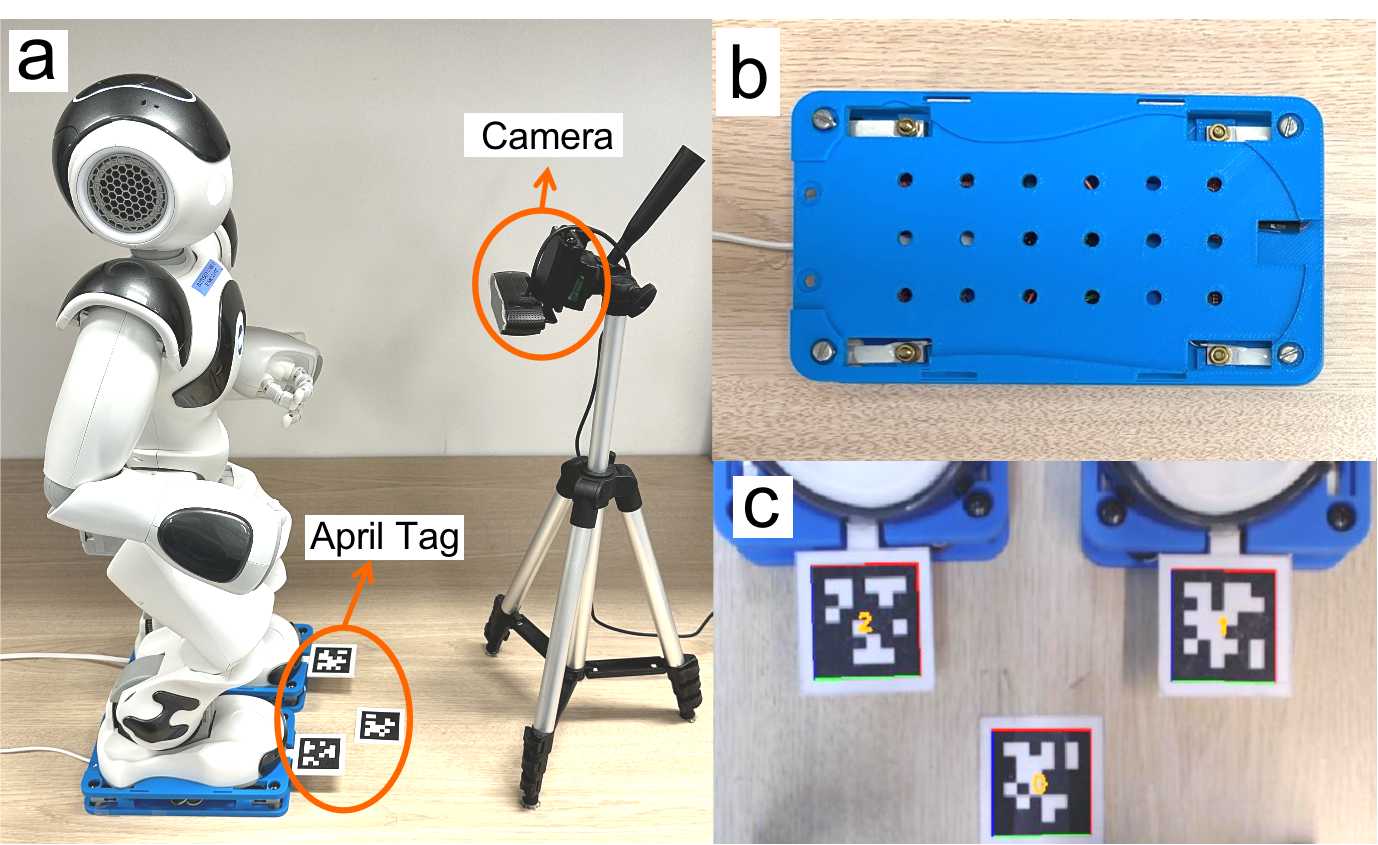}
\caption{(a) Overview of the experiment setup: a NAO robot, a pair of force-sensing shoes, three April tags, and a camera.  (b) Force-sensing shoe. (c) April tags scanned by the camera. }
\label{exp}
\end{figure}

The robot-sensor overview is shown in Fig. \ref{exp}a, which includes a Nao robot H25 V6, a pair of force-sensing shoes, three April tags, and a camera. The force-sensing shoes are the same as \cite{blueshoe}. We further attach an April tag to the bottom of the robot's foot to track foot positions (Fig. \ref{exp}b). April tags attached to the feet are called foot frames, and the one on the test bench is called the world frame.
\subsubsection{\bf{CoP Sensing Principle}}
Each force-sensing shoe has 4 load cells on the four corners of the shoe and measures the CoP. Through multiplying the force $f_i$ measured by each sensor by its corresponding 2D position ${\bf p}_i = [p_{ix}\;p_{iy}]^T$, and dividing by the sum of all the forces, the CoP, $\bf C$ in Eq. \ref{cost} is
\begin{align}
    {\bf C} = \sum_{i=1}^{4}f_{i}{\bf p}_{i}/ \sum_{i=1}^{4}f_{i}. \label{CoP_Equ}
\end{align}
Using all eight sensors, the same sensing principle that works for single support can also be used for double support.

\begin{figure}[t!]
\centering
\includegraphics[width=0.78\linewidth]{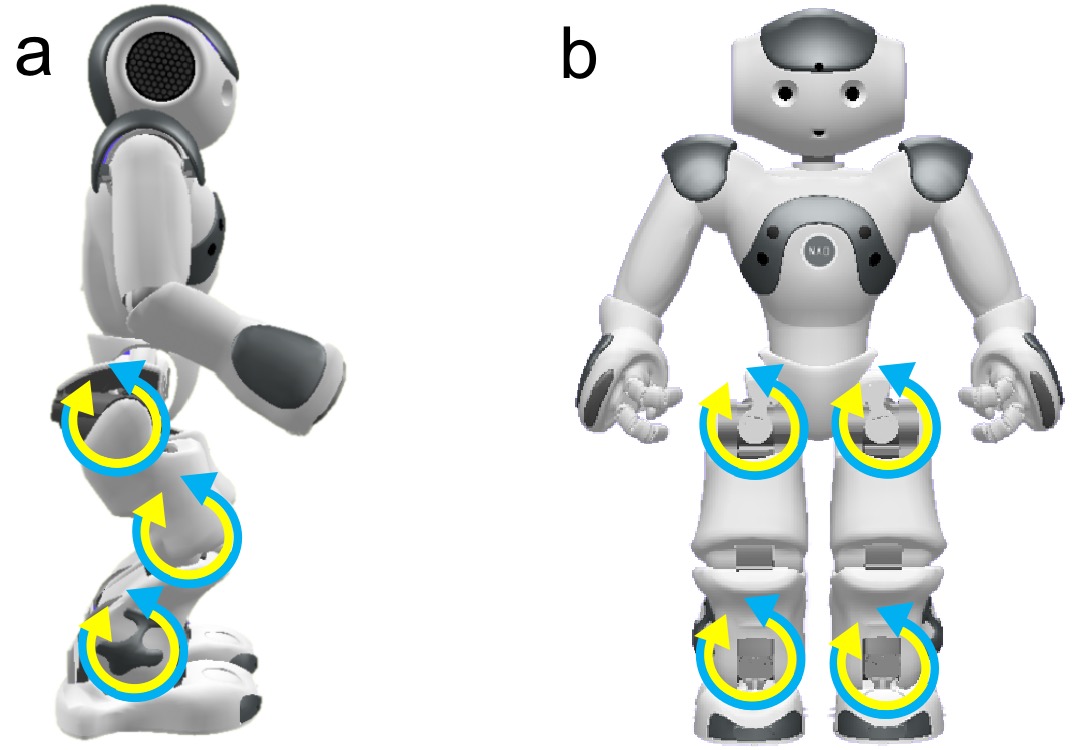}
\caption{Joints used for walking are marked with circled arrows. (a) Top middle and bottom are R(L)HipPitch, R(L)KneePitch and R(L)AnklePitch respectively. (b) Top and bottom are R(L)HipRoll and R(L)AnkleRoll respectively. R is right and L is left}
\label{NAO}
\end{figure}

\subsubsection{\bf{Foot Positions Measurement Principle}}\label{position}
There are four reference frames in the experiment: camera frame, world frame, foot frame, and floating base (Fig. \ref{gait}). Taking transition a $\rightarrow$ b as an example, to calculate ${\bf P}[j]$ in Eq. \ref{cost}, one needs to obtain the poses of the foot frames with respect to the floating base. From the camera and three April tags, one can directly get the transformation of the world frame and foot frames relative to the camera frame as ${\bf T}_0$, ${\bf T}_j$, where $j=l,r$ means left and right foot. Transfer the ${\bf T}_j$ into the world frame ${\bf T}_0$:
\begin{align}   
{{\bf T}_0^j} = {{\bf T}_0}^{-1} {\bf T}_j = \left[\begin{array}{cc}
{\text R_0^j} & {\bf p}_0^j \\
\bf{0}^{T} & 1
\end{array}\right]
\label{T}
\end{align}
where ${\text R_0^j} ={\text{R$_z$}} (\gamma_0^j) {\text{R$_y$}} (\beta_0^j){\text{R$_x$}} (\alpha_0^j)$\\
From Fig. \ref{gait}, the floating base is attached on the shoe (Fig. \ref{exp}a ). It is the same as the right foot frame except that it is always parallel to the test bench. That is, it will not rotate around $x$ or $y$-axis of the world frame with $\beta_0^r=0$, $\alpha_0^r=0 $. Therefore, from Eq. \ref{T}, the pose of the floating base with respect to the world frame is 
\begin{align}
{{\bf T}_0^{s}} = \left[\begin{array}{cc}
{\text {R$_z$}}(\gamma_0^r) & {\bf p}_0^r \\
\bf{0}^{T} & 1
\end{array}\right]
\end{align}
The pose of left foot frame with respect to the floating base is
\begin{align}
{{\bf T}_{s}^l} = {{\bf T}_0^{s}}^{-1} {\bf T}_0^l = 
\left[\begin{array}{cc}
{\text R_s^l} & {\bf p}_s^l \\
\bf{0}^{T} & 1
\end{array}\right]
\end{align}
where ${\text R_s^l} ={\text{R$_z$}} (\gamma_s^l) {\text{R$_y$}} (\beta_s^l){\text{R$_x$}} (\alpha_s^l)$\\
As a result, the ${\bf P}[l]$ in Eq. \ref{cost} is as
\begin{align}
{\bf P}[l] = [ {{\bf p}_s^l}^T \, \alpha_s^l \ \beta_s^l \  \gamma_s^l]^T
\end{align}
The pose of the right foot frame relative to the floating base and the sensors' positions can be derived by the same process.
\subsection{Optimization Algorithms}\label{optimization}
This section describes the optimization algorithm used to converge the robot's measured CoP and foot positions to their objectives. Simple gradient-based methods is used.

\begin{figure*}[t!]
\centering
\includegraphics[width=0.99\textwidth]{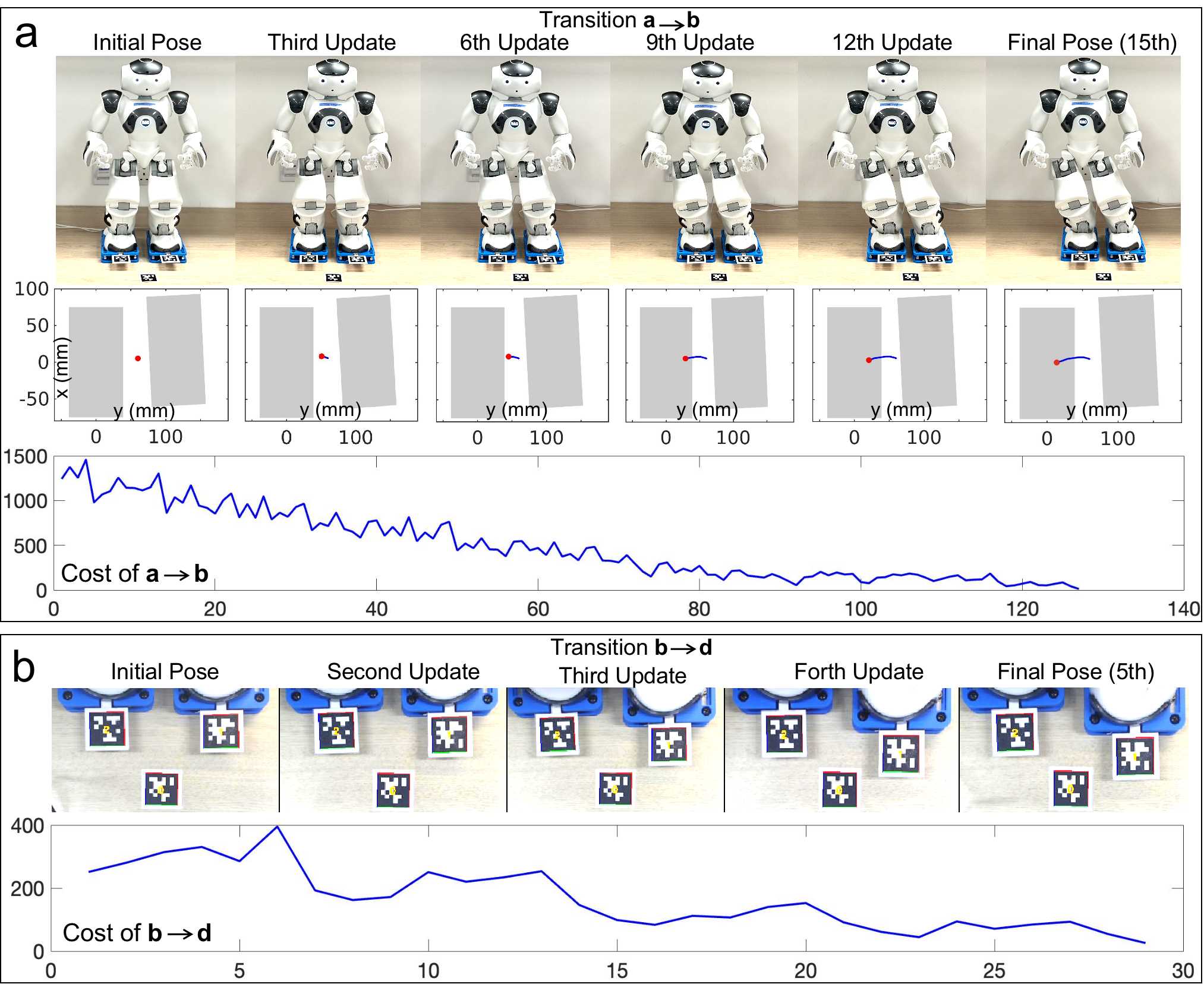}
\caption{Two example transitions. (a) Evolution of the robot’s motion, its CoP and cost for transition a $\rightarrow$ b. Grey rectangles are the robot shoes. (a) Evolution of the robot’s feet motion and cost for transition b $\rightarrow$ d}
\label{res}
\end{figure*}
{\bf Gradient Descent (GD)}
Given a multi-variable function $f({\bf q})$, the iterative law of the GD \cite{boyd2004convex} is 
\begin{equation}
{{{\bf q}}_{t+1}} = {{{\bf q}}_t} - \eta \nabla {f({{\bf q}}_t)}
\end{equation}
where $\nabla{f} = \frac{\partial \bf f}{\partial {\bf q}} $is the gradient of the objective function and $\eta$ is a step size that can be determined in different ways like Armijo rule \cite{armijo1966minimization}.

To approximate the derivatives in the optimization algorithm formula above, finite difference \cite{enwiki:1105295784} is used. In the literature, the finite difference method is often used due to its speed. Here it is used because there is not an explicit function whose derivative can be computed analytically. The function is empirical, resulting directly from sensor measurements and how their values are perturbed as the robot configuration is slightly varied.

\section{Experiment \romannumeral1}\label{expi}
\subsection{NAO Robot}
In this paper, the NAO V6 robot is used to test the general method. The mass is 5.3$kg$ and the height is 57$cm$. The robot has 25 degrees of freedom (DoFs). In this model-free biped walking framework, the five joints for each leg are considered in Fig. \ref{NAO}. 
\subsection{Model-Free Mapping of CoP Trajectory and Foot Placement to Joint Angles}\label{Map}
In the common humanoid robot, Pitch joints (Fig. \ref{NAO}a) are rotations around the $y$-axis, and Roll joints (Fig. \ref{NAO}b) are around the $x$-axis. Due to this structure, the robot can move the body's CoP horizontally along the $y$-axis by simply rotating four Roll joints of both legs, such as Fig. \ref{gait}a $\rightarrow$ b and d $\rightarrow$ e $\rightarrow$ f. For the same reason, it can lift one of its feet and step forward with the rotation of Pitch joints as Fig. \ref{gait}b $\rightarrow$ c $\rightarrow$ d. 

Taking transition Fig. \ref{gait}a $\rightarrow$ b as an example: assuming the joint vector in Fig. \ref{gait}a is ${\bf q}_0=[q_1\;q_2\;q_3\;q_4]^T$, NAO will rotate one of its joint forward for a certain angle as $f(q_1+\Delta q_1,q_2,q_3,q_4)$. This value can be used to calculate the gradient of $f({\bf q}_0)$. Each time the robot turns one or more joints and reaches a pose where the cost of calculating the derivatives can be found, it is called a \textit{search}. According to finite difference in \ref{optimization}, it takes 8 searches to find the gradient of $f({\bf q}_0)$ for an update. Then the joints are updated to the next state $\bf{q}_{01}$ by the optimization methods in \ref{optimization}. The robot continues this cycle until the value of the cost is less than a certain value or the difference between the costs of two consecutive updates is small enough. The same procedure is also taken in other transitions in Fig. \ref{gait}. The only difference is that the rotation joints are Pitch joints (Fig. \ref{NAO}a) in transition Fig. 1b $\rightarrow$ c $\rightarrow$ d.

\subsection{Demonstrations and Results} \label{Demo and Results}
To test the effectiveness of the method, we directly implemented it on a real robot without simulation. GD with a fixed step size is used to minimize the cost function in real-world experiments to present our method.

Transition examples are displayed in Fig. \ref{res}. Because the transitions d $\rightarrow$ e $\rightarrow$ f are the same as a $\rightarrow$ b with moving CoP horizontally along the $y$-axis, only examples of a $\rightarrow$ b $\rightarrow$ d are listed.  

There are two problems for NAO hardware when it supports its entire weight on a single leg: \textbf{1.} \textit{The HipRoll joint is unable to match the reference position. This is because the single support requires very high torque from this joint. As a result, the swinging leg does not leave the ground properly and will touch the ground prematurely, which destabilizes the robot} \cite{gouaillier2010omni}. \textbf{2.} \textit{For small-sized humanoid robots with under-powered motors, the single support generates excessive torque. This results in the NAO robot platform suffering from severe leg joint overheating issues} \cite{han2022watch}. To solve the first problem, the authors in \cite{gouaillier2010omni} design a trapezoid function feedback controller based on pre-planned joint trajectories. However, their method is model-based, which is not suitable for our model-free framework. As for the second problem, the authors in \cite{han2022watch} choose double support rather than single support for safety concerns. Therefore, we directly do the transition Fig. \ref{gait}b $\rightarrow$ d, without posture c. And a ${\bf C}_d$ closer to the center of double support is also chosen to reduce the joint torque. 

The results of generating gait trajectories with our model-free framework are displayed in Fig. \ref{res}. Fig. \ref{res}a shows that the CoP is getting closer to the target location and the cost function converges within a dozen of updates. Next, the robot can step forward with several times of updates Fig. \ref{res}b. There is no collision problem in the experiments. This is because the robot's joints only turn a small angle at each search. And the robot's desired left and right foot positions in the cost function have a certain distance between each other. The overall results demonstrate that our approach can generate a stable gait trajectory without the robot's inertial parameters, a kinematic model, reference joint signals, or a training process.  

\section{Experiment \romannumeral2}\label{expi2}

In our previous work \cite{han2022watch}, we utilized a robot model to plan whole-body trajectories and relied on modeled CoP and ground reaction force (GRF) as reference data to enable the robot to self-calibrate its shoe sensors. However, this model-based approach may not always be practical, especially when the robot is repaired or retrofitted with additional equipment, which may result in the loss of the robot model. To overcome this limitation, we integrated our model-free framework into the self-calibration process. This model-free self-calibration (MFSC) approach allows the robot to self-calibrate its shoe sensors without the need for a pre-existing model.

\subsection{Sensor Parameter Identification}
Our proposed method comprises two parts: 1. The robot follows planned CoP paths using our model-free method \ref{Methodology} when the shoe sensors are accurate. The measured CoP and GRF are saved as reference ground truth data. 2. When the sensors lose accuracy, the robot replays the previous trajectories, and the sensor parameters are optimized by reducing the difference between the reference data and the current measured data collected during the robot's movement. 

The load cell sensing principle is
\begin{align}
f = aS+b,
\label{affine}
\end{align}
where $S$ is the voltage output of the load cell, $a$ and $b$ are the constant coefficients.

\begin{figure}[t!]
\centering
\includegraphics[width=0.99\linewidth]{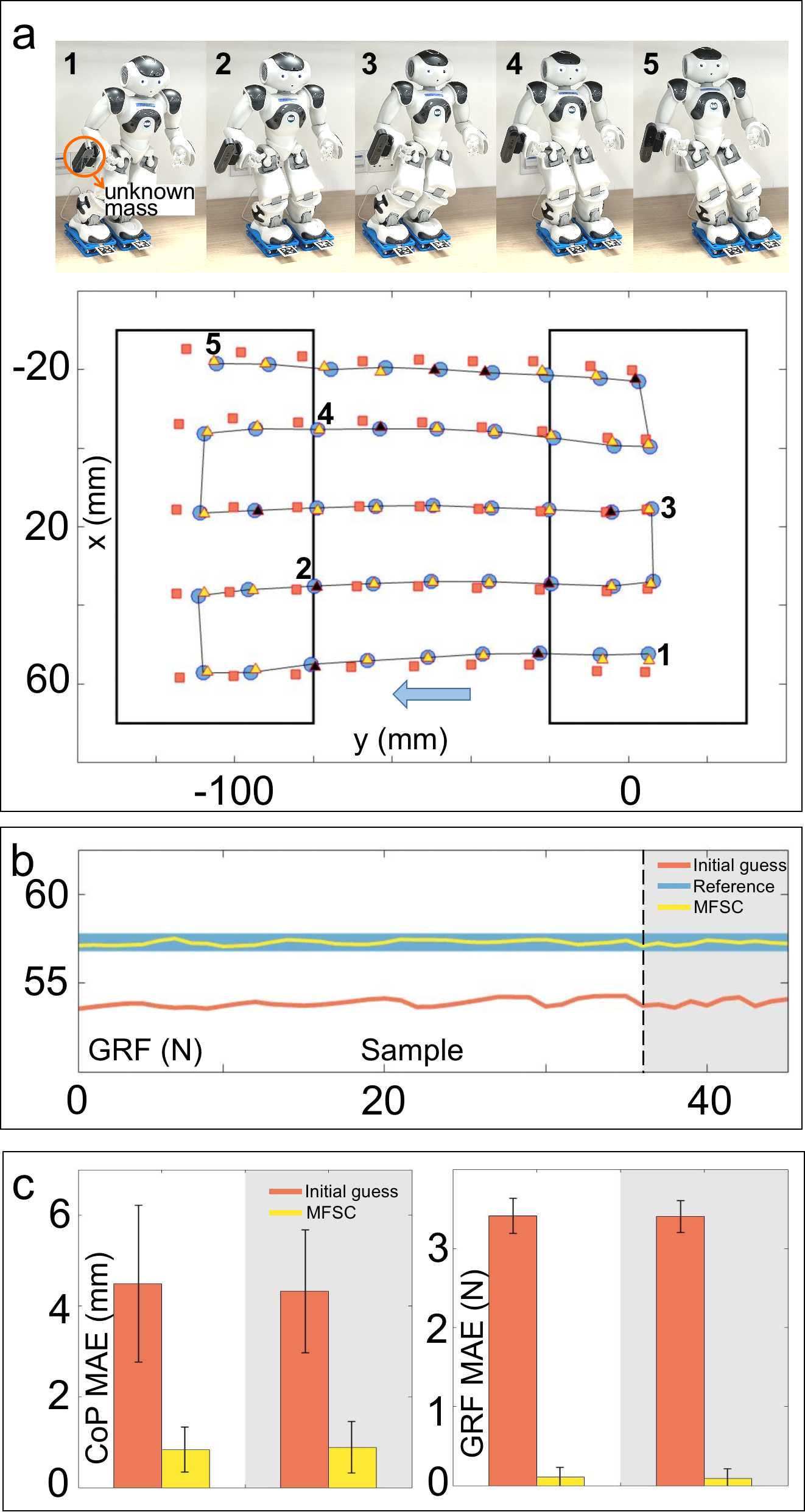}
\caption{Model-free self-calibration. (a) Top shows the example of the designed configurations. In bottom, black rectangles are the sensing polygon of each foot. Black line is the CoP trajectory and the blue arrow is the robot moving direction. The blue circles and red squares indicate the reference and initial guess CoPs. The yellow and black triangles represent the MFSC of the training and testing datasets. (b) and (c) Blue is the average of the reference data. Red and yellow show the measured data from the initial estimation and MFSC. The white and the grey backgrounds represent the training and testing datasets.}
\label{SC}
\end{figure}

To determine the load cells' parameters, we use nonlinear least squares (NLS) to minimize the error between the sensors' current raw measurements and their corresponding reference GRF and CoP values Fig. \ref{SC} a and b. The optimization is formulated as:
 \begin{align}\label{GRF_recovery} 
     \underset{{\bf \zeta}}{\text{argmin}}\quad J  = \sum_{k=1}^{N}(|n[k] - n_r|^{2}_{{\bf w_{n}}}+ \\ \nonumber
     ||{\bf c}[k] - {\bf c}_{r}[k]||^{2}_{{\bf w_{c}}} + ||{\bf f}[k]-{\bf f}_{r}[k]||^{2}_{{\bf w_{f}}}), 
\end{align}
where ${\bf \zeta} = [(a_{1},b_{1}),\hdots,(a_{8},b_{8})]$ are the optimization variables; $k$ is the number of training points; ${n}$ and ${\bf c}$ represent current GRF and CoP measurements; $ n_{r}$ and ${\bf c_{r}}$ are their reference data; ${\bf f}$ and ${\bf f}_r$ represent the measured force and reference force of each load cell. The weights for the cost terms are denoted as $w_{n}$, $w_{c}$, and $w_{f}$.
\subsection{Initial Guess of Sensor Parameters}
A reasonable initial guess is required for the NLS process. Given that all sensors in the shoes are of the same type, we use an equal initial estimation of $[a_0, b_0]$ for all sensors' parameters. We assume that the GRF and CoP measurements using this initial guess are approximately equivalent to their corresponding reference values, 
\begin{align}
    \begin{bmatrix}
    n_r\\
    {\bf c}_{r}
    \end{bmatrix} \approx 
    \begin{bmatrix}
    \sum_{i=1}^{8}(a_{0}S[i]+b_{0})\\
    \sum_{i=1}^{8}(a_{0}S[i]+b_{0}){\bf p}[i]/\sum_{i=1}^{8}(a_{0}S[i]+b_{0})
    \end{bmatrix}. 
\end{align}

$i$ represents the sensor index, and ${\bf p}$ is the sensor position obtained from the camera. To include more training samples, the equations can be rearranged as
\begin{align} \label{initial_guess_equ}
    \begin{bmatrix}
    \vdots\\
    {\bf c}_{r}[k]n_{r} \\
    \vdots
    \end{bmatrix} \approx 
    \begin{bmatrix}
    \vdots & \vdots\\
    \sum_{i=1}^{8}S[k][i]{\bf p}[k][i]&\sum_{i=1}^{8}{\bf p}[k][i]\\
    \vdots & \vdots
    \end{bmatrix} 
    \begin{bmatrix}
    a_{0}\\
    b_{0}
    \end{bmatrix}
\end{align}
where $k$ is the number of the training samples. The initial guess $(a_0,b_0)$ can therefore be resolved using least squares regression.
\subsection{Performance Evaluation}\label{error}
To assess the accuracy of the shoe measurements, mean absolute error (MAE) is used. The MAE of the GRF, $e_{n}$, is given by:
\begin{align}
&e_{n} = (\sum_{k=1}^{N}|n[k] - n_r[k]|)/N \label{eGRF},
\end{align}

The MAE of CoP, $e_{C}$, is defined as:
\begin{align}
&e_{c} = (\sum_{k=1}^{N}||{\bf c}[k] - {\bf c}_r[k]||)/N \label{eCoP}, 
\end{align}
\subsection{Demonstrations and Results}
To verify the efficacy of our MFSC approach, we compare the shoe measurements using the initial guess parameters with those using model-free self-calibrated parameters, as shown in Fig. \ref{SC}a and b. In this case, the camera only serves as an unknown mass, making the robot model uncertain. The corresponding MAE values for the two datasets are displayed in Fig. \ref{SC}c. The results indicate that the GRF (3.5 N MAE) and CoP (4.5 mm MAE) measurements through the initial estimation parameters are far from the ground truth(Fig. \ref{SC}c). On the contrary, the MAE of GRF and CoP measured by our MFSC approach are around 0.09 N and 0.8 mm respectively, which is much closer to the ground truth. Additionally, the performance of the training dataset (Fig. \ref{SC}c, white background) and testing dataset (grey background) does not differ significantly, indicating that there is no over-fitting. Overall, the results suggest that our model-free self-calibration method can effectively recover sensor measurements without relying on robot models, manual intervention or initial sensor data.

\section{Conclusion and Future work}
This research presents a novel model-free framework for quasi-static humanoid robot walking and self-calibrating foot force-sensing modules based only on sensor outputs. The proposed approach utilizes gradient descent to optimize the objective function that includes the measured CoP and positions of the robot's feet. We have directly implemented this framework on a real robot. The results show that it is capable of planning a steady walking trajectory without prior knowledge of the robot's inertial parameters or kinematic model. Although demonstrated on the NAO, this proprioceptive framework can theoretically be generalized to other robots. This method simply needs initial planning to generate one motion that can be saved for future usage. The framework utilizes rich sensor information, which can be integrated with other model-based techniques to explore unknown environments. Future work will involve adding filters to reduce sensor noise and implementing stochastic-based algorithms on various robots. With fast computation, we can also do dynamic proprioceptive motion by including the velocity and acceleration space in the search process.

\section{Acknowledgement}
This work was supported by NUS Startup    grants A-0009059-02-00, A-0009059-03-00, CDE Board account E-465-00-0009-01,  Tier 2 grant REBOT A-8000424-00-00, and National Research Foundation, Singapore, under its Medium Sized Centre Programme - Centre for Advanced Robotics Technology Innovation (CARTIN), sub award A-0009428-08-00.

\bibliographystyle{IEEEtran}
\bibliography{ICRA.bib}

\end{document}